\documentclass[sigconf]{acmart}

\renewcommand\footnotetextcopyrightpermission[1]{} 
\acmConference[KDD '26]{ACM SIGKDD Conference}{August 9--13, 2026}{Jeju, Korea}

\usepackage{booktabs, graphicx, xcolor, microtype, subcaption}
\usepackage{amsfonts, amsmath, amsthm} 
\usepackage{multirow, makecell, enumitem}
\usepackage{algorithm, algorithmic}
\usepackage{cleveref}
\usepackage{balance} 

\newtheorem{theorem}{Theorem}
\theoremstyle{definition}
\newtheorem{definition}[theorem]{Definition}

\usepackage{tikz}
\usetikzlibrary{arrows.meta, positioning}

\begin{document}

\title{Traceable Multi-Agent System for Knowledge-Based Forecasting}

\settopmatter{authorsperrow=4}

\author{Junhyeok Kang}
\authornote{These authors contributed equally to this research.}
\orcid{https://orcid.org/0009-0006-1569-7447}
\affiliation{
  \institution{LG AI Research}
  \city{Seoul}
  \country{Republic of Korea}
}
\email{junhyeok.kang@lgresearch.ai}

\author{Sangjun Han}
\authornotemark[1]
\orcid{https://orcid.org/0009-0004-6359-5061}
\affiliation{
  \institution{LG AI Research}
  \city{Seoul}
  \country{Republic of Korea}
}
\email{sj.han@lgresearch.ai}

\author{Hyeokjun Choe}
\authornotemark[1]
\affiliation{
  \institution{LG AI Research}
  \city{Seoul}
  \country{Republic of Korea}
}
\email{hyeokjun.choe@lgresearch.ai}

\author{Soonyoung Lee}
\affiliation{
  \institution{LG AI Research}
  \city{Seoul}
  \country{Republic of Korea}
}
\email{soonyoung.lee@lgresearch.ai}

\begin{teaserfigure}
    \centering
    \includegraphics[width=0.79\textwidth]{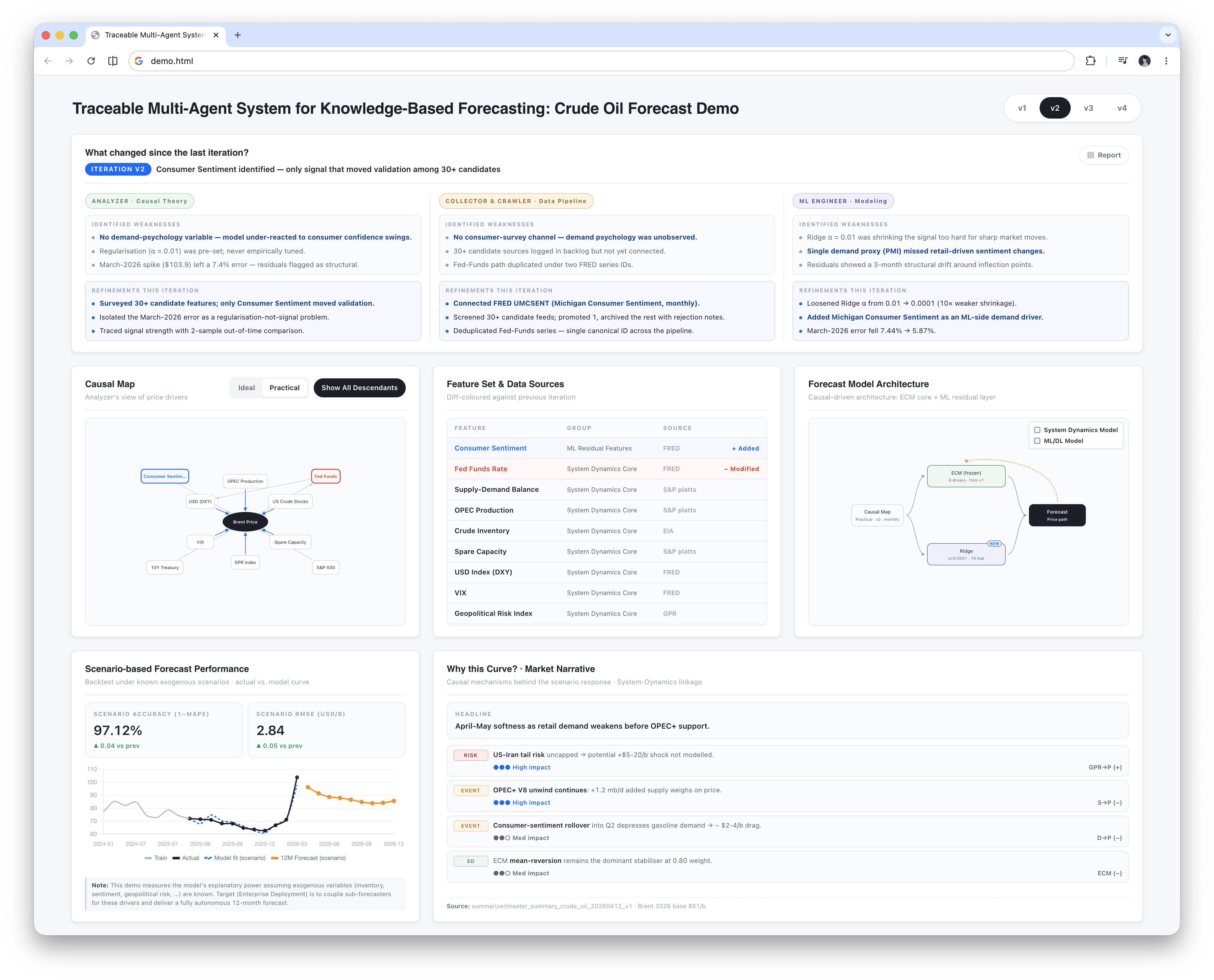}\vspace{-1.3em}
    \caption{TraceMAS demo interface for crude oil price forecasting. The interface allows users to inspect iteration-level agent revisions, causal maps, feature--data mappings, model architecture, scenario forecasts, and market narratives in a single view.}
    \label{fig:teaser}
    \vspace{0.8em}
\end{teaserfigure}

\begin{abstract}
    Enterprise forecasting increasingly relies on autonomous agents that interpret documents, search for data, generate code, and revise models. While this autonomy helps build adaptive forecasting pipelines, it also makes it difficult for practitioners to inspect why a forecast changed, which evidence supported the change, and how data and modeling choices were revised. We present \textbf{TraceMAS}, an interactive demo system for traceable multi-agent forecasting. TraceMAS organizes agent outputs around two causal-loop representations: an \textit{Ideal Causal Loop Diagram} (Ideal CLD), which captures key factors and their causal relations extracted from domain documents, and a \textit{Data-Grounded Causal Loop Diagram} (Data-Grounded CLD), which links those factors to internal variables, external data, or documented proxies. The Data-Grounded CLD guides feature construction and model design while preserving the connection between textual evidence, data choices, and model revisions. We demonstrate TraceMAS on crude oil price forecasting. The demo interface allows users to compare forecasting iterations, inspect agent-level revisions, explore causal maps, review feature--data mappings and model architecture, and connect scenario forecasts to market narratives. This demonstration shows how autonomous forecasting agents can retain flexibility while making the evidence-to-forecast process inspectable.
\end{abstract}



\maketitle


\section{Introduction}

Multi-agent systems are increasingly used to automate analytical workflows that require reasoning, tool use, and iterative revision~\cite{guo2024llmmas, wu2023autogen}. By assigning different roles to agents, these systems can decompose complex tasks, retrieve information, generate code, evaluate intermediate results, and revise earlier decisions. This autonomy is particularly useful when a workflow cannot be fully specified in advance, but must instead be constructed through interaction with heterogeneous sources of information. Enterprise forecasting is a natural setting for such agentic workflows. Forecasting pipelines often require practitioners to interpret market reports, identify factors affecting the target variable, find related indicators, construct features, and revise models as new evidence becomes available. A multi-agent system can distribute these steps across specialized agents, allowing textual domain knowledge and structured time-series data to be connected through an adaptive forecasting process.

The same autonomy, however, creates a tension in enterprise forecasting. Forecasts often support operational decisions such as material purchasing, inventory planning, and risk response~\cite{kang2025vardrop, trirat2024universal}. In such settings, practitioners need to understand not only the predicted value, but also the evidence and modeling choices behind it~\cite{arsenault2025survey}. When agents independently interpret documents, select variables, create features, and revise models, the rationale behind a forecast can become distributed across many intermediate outputs. As a result, a system may improve prediction accuracy while making it difficult to determine why the forecast changed, which evidence supported the change, or whether the change resulted from a reliable modeling decision.

This difficulty is especially important when forecasting relies on textual knowledge. Reports, expert analyses, and policy documents may describe conditions that are relevant to future movements of the target variable~\cite{williams2025context}, but such information must still be interpreted and connected to the data used by the forecasting model. For example, a report statement about supply constraints and future price pressure may lead the system to consider a supply-related indicator, create lagged features, or revise the model structure. If the forecast changes after these steps, practitioners need to inspect whether the change is supported by the original evidence, an appropriate data choice, and a reliable modeling process.

Existing approaches provide only partial support for this need. Text-informed forecasting methods can incorporate documents through embeddings, extracted signals, prompting, or cross-modal alignment, but they often hide how specific textual evidence affects variables, features, or models~\cite{wang2024newsforecast, jia2024gpt4mts, jin2024timellm, xu2024tgforecaster, liu2025calf}. Agent logs preserve more process information, but they are usually too unstructured to diagnose why a forecast changed~\cite{souza2025provagent}. For enterprise forecasting, traceability requires more than storing intermediate outputs; it requires a way to organize the links among textual evidence, data choices, feature construction, and model revisions~\cite{schlegel2025capturing}.

To address this challenge, we propose \textbf{TraceMAS}, a traceable multi-agent system for knowledge-grounded forecasting. TraceMAS preserves agent autonomy while organizing the intermediate outputs that connect textual knowledge to forecasting decisions. The system first extracts key factors affecting the target variable and document-supported causal hypotheses among them from domain documents, and synthesizes them into an \textit{Ideal Causal Loop Diagram} (Ideal CLD), which captures the causal structure suggested by textual knowledge before considering data availability. TraceMAS then reconstructs the Ideal CLD into a \textit{Data-Grounded Causal Loop Diagram} (Data-Grounded CLD) by linking factors to internal time-series variables, external data, or documented proxies. The Data-Grounded CLD guides variable selection, feature construction, and model design while preserving how textual evidence is connected to the forecasting pipeline. By maintaining versioned CLDs across iterations, TraceMAS allows practitioners to inspect which textual evidence contributed to causal assumptions, how those assumptions were linked to data, which features were constructed, and how model revisions changed the forecast.

This demo makes the following contributions:
\begin{itemize}
    \item We present TraceMAS, an interactive multi-agent forecasting system that exposes how domain documents, data sources, features, model revisions, and forecasts are connected.

    \item We introduce Ideal CLD and Data-Grounded CLD as shared representations for tracing how textual evidence is transformed into causal assumptions, data choices, features, and model components.

    \item We demonstrate TraceMAS on crude oil price forecasting, showing how versioned CLDs help inspect the transformation from market-report knowledge to forecast revisions.
\end{itemize}

\section{Proposed Framework: TraceMAS}

TraceMAS builds forecasting models by connecting textual domain knowledge with structured time-series data. The system allows agents to explore and revise the forecasting pipeline autonomously, while organizing their outputs into traceable intermediate representations. Rather than imposing a rigid linear procedure, TraceMAS records agent-generated knowledge, data mappings, features, review signals, and model outputs through a CLD-centered design. The recorded artifacts form a versioned trace, where each iteration stores the current CLDs, data mappings, feature specifications, reviewer warnings, model configuration, and forecast outputs.

\subsection{Workflow Overview}

\begin{figure*}[t]
    \centering
    \resizebox{\textwidth}{!}{%
    \begin{tikzpicture}[
        font=\small,
        box/.style={
            draw,
            rounded corners=3pt,
            align=center,
            minimum width=3.0cm,
            minimum height=1.0cm
        },
        arrow/.style={-Latex, thick}
    ]

    \node[box] (docs) at (0, 2) {Domain Documents\\(.pdf, .docx, ...)};
    \node[box] (ideal) at (4, 2) {Ideal\\Causal Loop Diagram};

    \node[box] (db) at (0, 0) {Enterprise\\Time-series Database};
    \node[box] (ext) at (4, 0) {External Data Source};

    \node[box] (grounded) at (8.2, 1) {Data-Grounded\\Causal Loop Diagram};
    \node[box] (model) at (12.2, 1) {Forecasting Model};
    \node[box] (forecast) at (16, 1) {Forecasts};

    \draw[arrow] (docs) -- (ideal);
    \draw[arrow] (ideal.east) -- ++(0.5,0) |- (grounded.west);

    \draw[arrow] (db.east) -- ++(0.5,0) |- (grounded.west);
    \draw[arrow] (ext.east) -- ++(0.5,0) |- (grounded.west);

    \draw[arrow] (grounded) -- (model);
    \draw[arrow] (model) -- (forecast);

    \end{tikzpicture}%
    }
    \caption{Overview of the TraceMAS workflow. Domain documents are used to construct an Ideal CLD. The Ideal CLD is then linked to enterprise time-series data and external data sources to form a Data-Grounded CLD, which guides forecasting model construction and final prediction.}
    \label{fig:workflow}
\end{figure*}
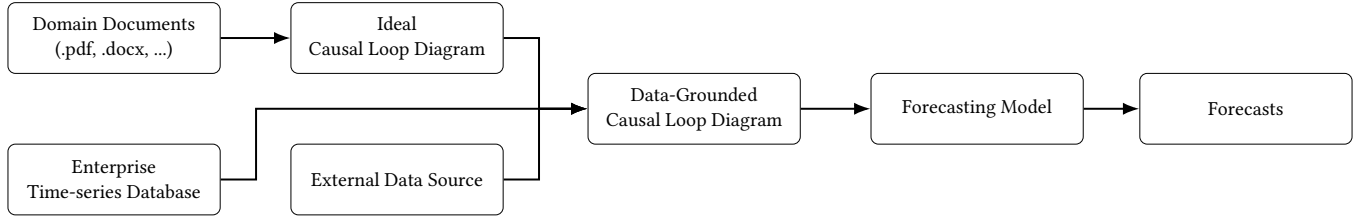

The TraceMAS workflow consists of document-based knowledge extraction, CLD construction, data mapping, model construction, evaluation, and iterative revision. Figure~\ref{fig:workflow} summarizes the workflow from raw inputs to forecasting outputs. First, the \textit{Domain Analyst} extracts key factors that influence the target variable and identifies directional hypotheses suggested by the documents~\cite{hosseinichimeh2024text}. These hypotheses are not treated as keyword co-occurrences; they record directional influences between factors, often with indicated time delays. For example, the sentence ``if crude oil supply becomes constrained, prices may rise after three months'' is recorded as a delayed directional hypothesis from supply shortage to price increase.

The \textit{Causal Analyst} aggregates extracted factors and document-supported hypotheses into an Ideal CLD, which represents the causal structure suggested by documents before considering data availability. The \textit{Data Engineer} searches the internal time-series database for variables or historical patterns related to each factor, reflecting the importance of retrieving relevant time-series information for feature selection in forecasting~\cite{kang2026channel}. If no suitable internal variable exists, the \textit{Crawler} searches for external data sources or proxy candidates. The \textit{Coordinator} then synthesizes the Ideal CLD, data mappings, proxy candidates, model results, and review signals to update the Data-Grounded CLD.

The \textit{Model Engineer} uses the Data-Grounded CLD to construct features and forecasting models. Factors become candidate features, while document-supported hypotheses inform lag choices, interactions, and structural components. The \textit{Risk Reviewer} checks generated data, code, mappings, and model outputs for leakage, timestamp misalignment, inconsistent units, missing sources, and hard-coded values. Across iterations, changes to factors, hypotheses, proxies, features, review signals, and model components are recorded as versioned updates, allowing users to compare how agent decisions propagate through the forecasting pipeline.

\subsection{CLD-Based Shared Representation}

Causal loop diagrams are widely used in system dynamics to represent feedback structures and directional relations in complex systems~\cite{sterman2000business}. TraceMAS uses CLDs as shared artifacts for organizing the forecasting process. A CLD summarizes document-derived factors, directional hypotheses, data links, feature transformations, and model components used in the current iteration. Nodes denote key factors that may influence the target variable, while edges denote document-supported directional hypotheses among factors. In crude oil price forecasting, for instance, inventory, production outlook, demand outlook, and geopolitical risk may appear as nodes, with edges representing their possible influence on supply pressure and price movement.

\begin{definition}[\textit{Ideal CLD}]
An Ideal CLD is a directed graph $G_I=(V_I,E_I)$. Each node $v \in V_I$ denotes a key factor extracted from textual evidence, and each edge $e=(v_i,v_j,\tau) \in E_I$ denotes a document-supported directional hypothesis from factor $v_i$ to factor $v_j$ after a time delay $\tau$. The Ideal CLD captures the causal structure suggested by documents before considering whether each factor is observable in available data.
\end{definition}

\begin{definition}[\textit{Data-Grounded CLD}]
A Data-Grounded CLD is defined as $G_D=(V_D,E_D,M)$. The nodes $V_D$ and edges $E_D$ are derived from the Ideal CLD by selecting, merging, replacing, or extending factors and hypotheses according to data availability and proxy feasibility. The mapping function $M$ links each node $v \in V_D$ to an internal time series, an externally collected time series, or a documented proxy.
\end{definition}

The Data-Grounded CLD is the main inspectable representation in the demo. It shows which document-derived factors are used for forecasting, how they are linked to data, and how they guide feature construction, lag selection, interactions, and model components. Because the Data-Grounded CLD is updated across iterations, users can compare versions to inspect which factors were added, which proxies were replaced, and which modeling choices changed. Thus, the Data-Grounded CLD is not a post-hoc explanation; it is a shared working artifact used during forecast construction and revision.

\subsection{Role-Based Agents}

TraceMAS consists of a main agent, \textit{Coordinator}, and six subagents: \textit{Domain Analyst}, \textit{Causal Analyst}, \textit{Data Engineer}, \textit{Crawler}, \textit{Risk Reviewer}, and \textit{Model Engineer}. These agents are not merely stages in a fixed linear pipeline. They are roles that update shared intermediate outputs from different perspectives.

\begin{enumerate}

    \item \textit{Domain Analyst} extracts key factors and document-supported directional hypotheses from domain documents, preserving source evidence, directionality, temporal expressions, and uncertainty.

    \item \textit{Causal Analyst} synthesizes extracted factors and hypotheses into an Ideal CLD and updates the causal structure as new factors, relations, or revisions are proposed.

    \item \textit{Data Engineer} searches the internal time-series database for variables corresponding to Data-Grounded CLD nodes and checks coverage, frequency, units, and missingness.

    \item \textit{Crawler} retrieves external data or proxy candidates for factors that are not directly available in the internal database, recording their source, coverage, and limitations.

    \item \textit{Risk Reviewer} continuously checks generated data, code, mappings, and model outputs for leakage, timestamp misalignment, inconsistent units, missing sources, and hard-coded values.

    \item \textit{Model Engineer} builds CLD-guided features and forecasting models, using document-supported hypotheses to inform lags, interactions, and structural components.

    \item \textit{Coordinator} synthesizes the Ideal CLD, data mappings, proxy candidates, model results, and review signals to update the Data-Grounded CLD and issue revision requests across iterations.

\end{enumerate}

\begin{table*}[t]
    \centering
    \footnotesize
    \caption{Representative outputs from the TraceMAS demo. Each row shows how a document-derived hypothesis is connected to data sources and then used as a model input.}
    \label{tab:causal_analyst_output}
    \begin{tabular}{p{0.18\textwidth} p{0.29\textwidth} p{0.34\textwidth} p{0.11\textwidth}}
    \toprule
    \textbf{Factor} & \textbf{Document-Derived Hypothesis} & \textbf{Data Used} & \textbf{Model Input} \\
    \midrule
    Supply--Demand Balance
    & Market surplus $\rightarrow$ downward price pressure.
    & Production, demand, and inventory series.
    & Balance feature. \\

    OPEC+ Production Policy
    & Production unwinding $\rightarrow$ higher supply pressure.
    & OPEC+ monthly production data.
    & Policy feature. \\

    China Demand and Stockpiling
    & Demand recovery or stockpiling $\rightarrow$ price support.
    & China imports, PMI, and stockpiling proxies.
    & Demand lag feature. \\

    Geopolitical Risk
    & Conflict or sanctions $\rightarrow$ supply disruption risk.
    & GPR index, news-based tension proxies, or sanctions series.
    & Risk feature. \\
    \bottomrule
    \end{tabular}
\end{table*}

\section{Demonstration: Crude Oil Price Forecasting}

We demonstrate TraceMAS through an interactive crude oil price forecasting interface. The demo focuses on process traceability: how document-derived market knowledge is converted into data choices, feature transformations, model revisions, and forecast narratives that users can inspect across iterations. As shown in Figure~\ref{fig:teaser}, users can select a forecasting iteration and inspect agent-level revision summaries, a causal map, feature--data mappings, model architecture, scenario-based forecast performance, and a market narrative in a single view.

\subsection{Interface and Inspection Workflow}

The interface supports an iteration-level inspection workflow. Users first select a forecasting iteration and review what changed since the previous version through agent-level revision summaries. They can then inspect the causal map, compare document-derived factors with their data-linked counterparts, review feature--data mappings, and examine the corresponding model architecture. The scenario forecast and market narrative panels summarize how the current model behavior is explained in terms of market conditions, selected variables, and revision history.

The interface also exposes reviewer warnings generated during each iteration. Users can inspect whether a proxy has insufficient temporal coverage, whether a feature uses data unavailable at the forecast time, whether a data source is missing for a claimed relation, or whether generated code contains hard-coded values. These components allow users to follow how textual evidence becomes causal assumptions, mapped variables, feature transformations, model components, and forecast narratives. Selecting a factor, feature, or reviewer warning opens the corresponding evidence, data mapping, feature transformation, or revision rationale. This allows users to distinguish forecast changes caused by new evidence, data mapping decisions, feature revisions, or reviewer-flagged risks.

\subsection{From Documents to Forecast Revisions}

In each iteration, TraceMAS analyzes crude oil market reports and organizes price drivers into direct, semi-direct, and indirect factors. The Domain Analyst extracts key factors and document-supported hypotheses, while the Causal Analyst synthesizes them into an Ideal CLD. The Data Engineer and Crawler identify which factors can be linked to variables in the enterprise time-series database, external data sources, or documented proxies, and the Coordinator updates the Data-Grounded CLD for model construction. Table~\ref{tab:causal_analyst_output} shows representative outputs generated in the demo. Each row corresponds to a traceable transformation from a document-derived hypothesis to data sources and then to a model input.

After the Data-Grounded CLD is constructed, the Model Engineer builds CLD-guided features and forecasting models. Direct factors such as supply--demand balance become market-state features, while relations with temporal delay guide lagged features. Factors unavailable in the internal database, such as geopolitical risk or demand outlook, are represented through external indices, news-based proxies, or derived indicators. Across iterations, TraceMAS revises both the causal map and the corresponding model inputs: newly added factors may trigger proxy search and feature construction, while weak or insufficiently covered proxies can be flagged by the Risk Reviewer and replaced in later iterations. The interface reports validation metrics only to help users compare iterations and inspect how CLD updates, proxy choices, and feature revisions affect model behavior, rather than to claim benchmark-level forecasting performance.

\subsection{Inspecting the Evidence-to-Forecast Path}

Users can compare the Ideal CLD and Data-Grounded CLD across iterations to inspect which factors were added, which proxies were replaced, and which relation directions or lag assumptions were revised. They can also select a factor or relation and trace it back to the source document, the mapped variable or proxy, the feature transformation, and the model component that uses it. For instance, a demand-related relation can be inspected as a path from market commentary on demand recovery or stockpiling, to a demand factor in the CLD, to import or proxy series, and finally to lagged demand features used by the forecasting model. This demonstration shows how TraceMAS preserves the intermediate outputs needed to inspect how the forecasting pipeline evolves. Figures~\ref{fig:initial_ideal_cld} and~\ref{fig:initial_data_grounded_cld} in Appendix~\ref{app:initial_cld_examples} show the initial CLD artifacts generated by TraceMAS.

\section{Discussion and Conclusion}

This paper presents TraceMAS, an interactive demo system for traceable enterprise forecasting with LLM-based agents. Instead of incorporating text only through latent embeddings, TraceMAS extracts key factors and document-supported hypotheses from domain documents, organizes them into an Ideal CLD, links them to real data through a Data-Grounded CLD, and constructs CLD-guided forecasting models. The crude oil price demo shows how documents, data, variables, model components, forecasts, and market narratives remain inspectable across iterative agent revisions. The demo illustrates that traceability in agentic forecasting can be supported not only by storing logs, but by exposing structured, versioned artifacts that users can inspect during the forecasting process. The core contribution of TraceMAS is to preserve agent autonomy while keeping the transformation from domain knowledge to forecasting decisions explicit and comparable across iterations.

\newpage
\bibliographystyle{ACM-Reference-Format}
\balance 
\bibliography{6-Reference}

\clearpage
\onecolumn
\appendix

\section{Initial CLD Examples in Crude Oil Price Forecasting}
\label{app:initial_cld_examples}

\begin{figure}[h]
    \centering
    \includegraphics[
        width=1.1\textwidth,
        trim=2cm 5cm 0cm 5.0cm,
        clip
    ]{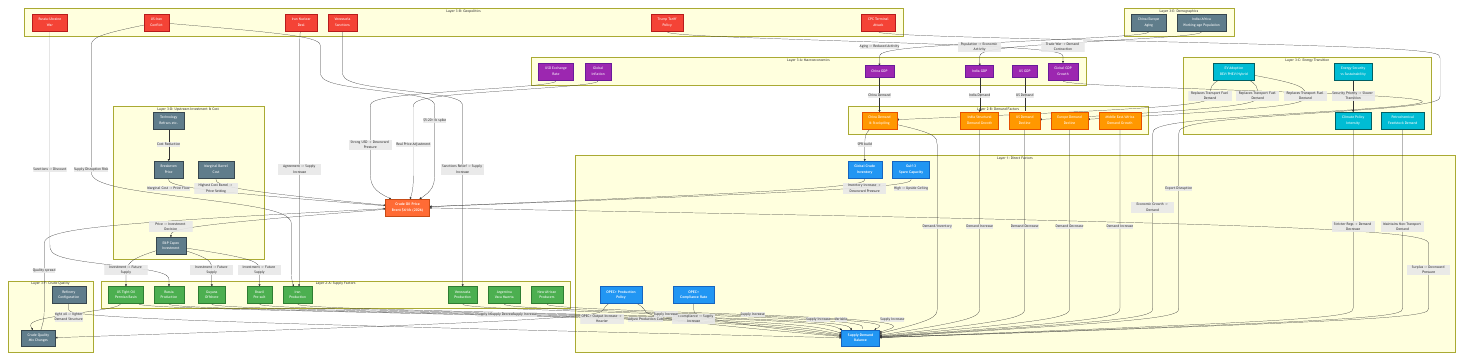}
    \vspace{-10em}
    \caption{Initial Ideal CLD. The diagram summarizes document-derived factors and directional hypotheses extracted from crude oil market reports before considering data availability.}
    \label{fig:initial_ideal_cld}
\end{figure}

\begin{figure}[h]
    \centering
    \includegraphics[
        width=1.1\textwidth,
        trim=2.5cm 5cm 0cm 2.0cm,
        clip
    ]{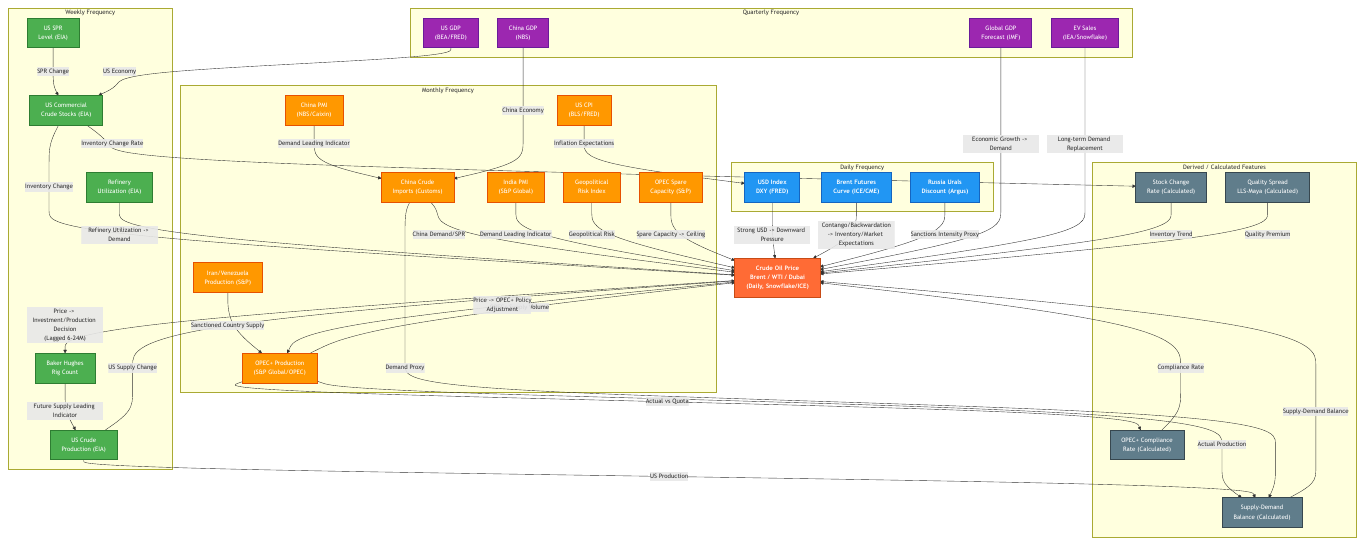}
    \vspace{-9em}
    \caption{Initial Data-Grounded CLD. The diagram links document-derived factors to available internal time-series variables, external data sources, or documented proxies used for forecasting.}
    \label{fig:initial_data_grounded_cld}
\end{figure}

Figures~\ref{fig:initial_ideal_cld} and~\ref{fig:initial_data_grounded_cld} show the initial CLD artifacts generated by TraceMAS in the crude oil price forecasting demo. These figures illustrate the distinction between the document-derived causal structure and its data-grounded counterpart. The Ideal CLD represents factors and directional hypotheses extracted from crude oil market reports before considering whether each factor is directly observable in available data. It includes broad domain concepts such as supply--demand balance, OPEC+ production policy, inventory, spare capacity, demand-side factors, macroeconomic conditions, geopolitical risk, energy transition, and upstream investment.

The contrast between the two CLDs shows how TraceMAS separates two forms of traceability. The Ideal CLD preserves the reasoning structure suggested by textual evidence, while the Data-Grounded CLD records how that reasoning structure is linked to observable variables, external data sources, or documented proxies for forecasting. This separation allows users to inspect whether a forecast revision came from a change in document-derived assumptions, a change in data availability, a proxy replacement, or a feature transformation. In later iterations, the same representation is updated with model evidence and reviewer feedback, enabling users to compare CLD versions and trace how the forecasting pipeline evolves.

\end{document}